\def\year{2021}\relax
\documentclass[letterpaper]{article} 
\usepackage{aaai21}  
\usepackage{times}  
\usepackage{helvet} 
\usepackage{courier}  
\usepackage[hyphens]{url}  
\usepackage{graphicx} 
\usepackage{natbib}  
\usepackage{caption} 
\usepackage[switch]{lineno}  %
\newcommand{\etal}{\textit{et al}.~}

\newcommand{\ieno}{\textit{i}.\textit{e}.}

\newcommand{\egno}{\textit{e}.\textit{g}.} 

\usepackage{array}
\newcolumntype{C}[1]{>{\centering\arraybackslash}m{#1}}
\newcolumntype{R}[1]{>{\raggedleft\arraybackslash}m{#1}}
\newcolumntype{P}[1]{>{\raggedright\arraybackslash}p{#1}}
\newcolumntype{M}[1]{>{\centering\arraybackslash}m{#1}}

\newcommand{\tcb}{\textcolor{black}}
\usepackage{algorithm}

\usepackage{algpseudocode}
\usepackage{amsmath}
\usepackage{booktabs}       
\usepackage{multirow}
\usepackage{comment}

\usepackage{subfigure}

\usepackage{enumitem}
\usepackage{xcolor}
\usepackage{amsfonts}

\title{Exploiting Sample Uncertainty for Domain Adaptive Person Re-Identification}

\author {
        Kecheng Zheng\textsuperscript{\rm 1}\thanks{This work was done when Kecheng Zheng was an intern at MSRA.},
        Cuiling Lan\textsuperscript{\rm 2}\thanks{Corresponding Author},
        Wenjun Zeng\textsuperscript{\rm 2},
        Zhizheng Zhang\textsuperscript{\rm 1},
        Zheng-Jun Zha\textsuperscript{\rm 1}$^{\dag}$ \\
}
\affiliations {
    \textsuperscript{\rm 1} University of Science and Technology of China \\
    \textsuperscript{\rm 2} Microsoft Research Asia \\
    \{zkcys001,zhizheng\}@mail.ustc.edu.cn, \{culan, wezeng\}@microsoft.com, zhazj@ustc.edu.cn
}
\begin{document}
	\maketitle
	
	\begin{abstract}
		Many unsupervised domain adaptive (UDA) person re-identification (ReID) approaches combine clustering-based pseudo-label prediction with feature fine-tuning. However, because of domain gap, the pseudo-labels are not always reliable and there are noisy/incorrect labels. This would mislead the feature representation learning and deteriorate the performance. In this paper, we propose to estimate and exploit the credibility of the assigned pseudo-label of each sample to alleviate the influence of noisy labels, by suppressing the contribution of noisy samples. We build our baseline framework using the mean teacher method together with an additional contrastive loss. We have observed that a sample with a wrong pseudo-label through clustering in general has a weaker consistency between the output of the mean teacher model and the student model. Based on this finding, we propose to exploit the uncertainty (measured by consistency levels) to evaluate the reliability of the pseudo-label of a sample and incorporate the uncertainty to re-weight its contribution within various ReID losses, including the identity (ID) classification loss per sample, the triplet loss, and the contrastive loss. Our uncertainty-guided optimization brings significant improvement and achieves the state-of-the-art performance on benchmark datasets.
	\end{abstract}

	\begin{figure}[th]
		\centering
		\includegraphics[width=1.0\linewidth]{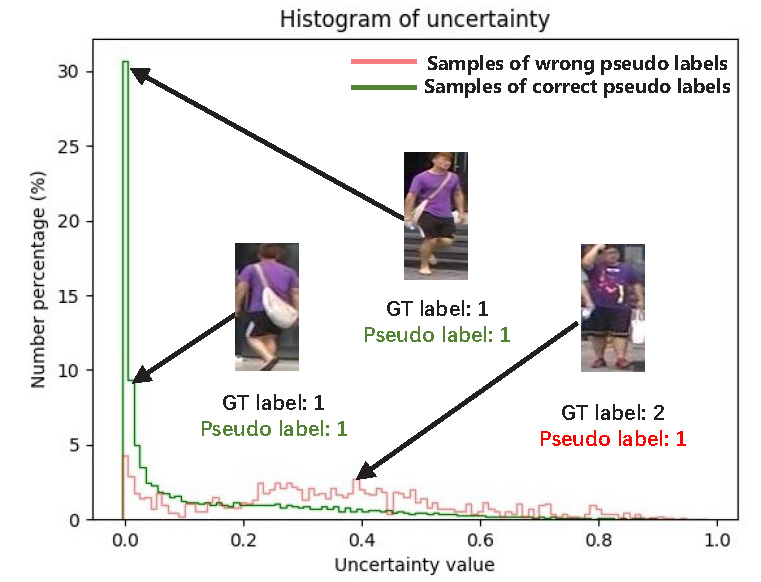}
		\caption{Observations on the relations between the correctness of pseudo labels and the uncertainty (which we measure by the inconsistency level of the output features of two models, \ieno, the student model and the teacher model based on the mean teacher method~\cite{tarvainen2017mean} for the target domain samples (obtained from Duke$\to$Market). We found the uncertainty for samples with wrong/noisy pseudo labels (red curve) is usually larger than those (green curve) with correct/clean pseudo labels.}

		\label{fig:intro}
	\end{figure}

	\section{Introduction}
	Person re-identification (ReID) is an important task that matches person images across times/spaces/cameras, which has many applications such as people tracking in smart retail, image retrieval for finding lost children. Existing approaches achieve remarkable performance when the training and testing data are from the same dataset/domain. But they usually fail to generalize well to other datasets where there are domain gaps~\cite{ge2020mutual}. To address this practical problem, unsupervised domain adaptive (UDA) person ReID attracts much attention for both the academic and industrial communities, where labeled source domain and unlabeled target domain data are exploited for training.

	Typical UDA person ReID approaches~\cite{ge2020mutual,zhai2020adcluster,zhong2019invariance,2020Hierarchical,song2020unsupervised} include three steps: feature pre-training with labeled source domain data, clustering-based pseudo-label prediction for the target domain data, and feature representation learning/fine-tuning with the pseudo-labels. The last two steps are usually iteratively conducted to promote each other. However, the pseudo-labels obtained/assigned through clustering usually contain noisy (wrong) labels due to the divergence/domain gap between the source and target data, and the imperfect results of the clustering algorithm. Such noisy labels would mislead the feature learning and harm the domain adaptation performance. Thus, \emph{alleviating the negative effects of those samples with unreliable/noisy pseudo labels is important for the success of domain adaptation}.



	The challenge lies in 1) how to identify samples that are prone to have noisy pseudo labels; 2) how to alleviate their negative effects during the optimization. In this paper, to answer the first question, we have observed abundant samples and analyzed the relationship between the characteristics of the samples and the correctness of pseudo labels. Based on the theory on uncertainty~\cite{kendall2017uncertainties}, a model has uncertainty on its prediction of an input sample. Here, we measure the inconsistency level of the output features of two models (the student model and the teacher model based on the mean teacher method~\cite{tarvainen2017mean}) and take it as the estimated uncertainty of a target domain sample. \tcb{As shown in Fig.~\ref{fig:intro}, we observe the distribution of the uncertainty (inconsistency levels) for correct/clean pseudo labels and wrong pseudo labels. We found that the uncertainty values for the samples with wrong pseudo labels are usually larger} than those with correct pseudo labels. This motivates us to estimate and exploit the uncertainty of samples to alleviate the negative effects of noisy pseudo labels, enabling effective domain adaptation. We answer the second question by carefully incorporating the uncertainty of samples into classification loss, triplet loss, and contrastive loss, respectively.

	We summarize our main contributions as follows:

\begin{itemize}[leftmargin=*,noitemsep,nolistsep]

\item  We propose a network named Uncertainty-guided Noise Resilient Network (UNRN) to explore the credibility of the predicted pseudo labels of target domain samples for effective domain adaptive person ReID.

\item  We develop an uncertainty estimation strategy by calculating the inconsistency of two models in terms of their predicted soft multilabels.

\item We incorporate the uncertainty of samples to the ID classification loss, triplet loss, and contrastive loss through re-weighting to alleviate the negative influence of noisy pseudo labels.

\end{itemize}

Extensive experiments demonstrate the effectiveness of our framework and the designed components on unsupervised person ReID benchmark datasets. Our scheme achieves the state-of-the-art performance on all the benchmark datasets.

	\section{Related Work}

	\subsection{Unsupervised Domain Adaptive Person ReID}
	Existing unsupervised domain adaptive person ReID approaches can be grouped into three main categories.

	\noindent\textbf{Clustering-based methods} in general generate hard or soft pseudo labels based on clustering results and then fine-tune/train the models based on the pseudo labels ~\cite{fan2018unsupervised,zhang2019self,yang2019selfsimilarity,ge2020mutual,yu2019unsupervised,zhong2019invariance, song2020unsupervised, jin2020global}. They are widely used due to their superior performance. PUL~\cite{fan2018unsupervised} iteratively obtains hard clustering labels and trains the model in self-training manner. SSG~\cite{yang2019selfsimilarity} exploits the potential similarity for the global body and local body parts, respectively, to build multiple independent clusters. These clusters are then assigned with labels to supervise the training. PAST~\cite{zhang2019self} optimizes the network with triplet-based loss function (to capture the local structure of target-domain data points) and classification loss by appending a classification layer (to use global information about the data distribution) based on clustering results.

	Pseudo label noise caused by unsupervised clustering is always an obstacle to the self-training. Such noisy labels would mislead the feature learning and impede the achievement of high performance. Recently, some methods introduce mutual learning among two/three collaborative networks to mutually exploit the refined soft pseudo labels of the peer networks as supervision \cite{ge2020mutual,zhai2020multiple}. To suppress the noises in the pseudo labels, NRMT \cite{zhao2020unsupervised} maintains two networks during training to perform collaborative clustering and mutual instance selection, which reduces the fitting to noisy instances by using the mutual supervision and the reliable instance selection. These approaches need mutual learning of two or more networks and are somewhat complicated. Besides, the selection of reliable instances in NRMT~\cite{zhao2020unsupervised} is a hard rather than a soft selection, which requires a careful determination of threshold parameters and may lose the chance of exploiting useful information of the abandoned samples.

	Different from the above works, in this paper, we propose a simple yet effective framework which estimates the reliability of the pseudo labels through uncertainty estimation and softly exploit them in the ReID losses to alleviate the negative effects of noise-prone samples. Note that we do not need two networks for mutual learning. We build our framework based on the mean teacher method, which maintains a temporally averaged model of the base network during the training, to facilitate the estimation of the uncertainty of each target domain sample.


	\noindent\textbf{Domain translation-based methods}  \cite{deng2018image,huang2020real,wei2018person,ge2020structured} transfer source domain labeled images to the style of the target domain images and use these transferred images and the inherited ground-truth labels to fine-tune the model. However, the quality of translated images is still not very satisfactory which hinders the advancement of these approaches.

	\noindent\textbf{Memory bank based methods} have been widely used for unsupervised representation learning which facilitates the introduction of contrastive loss~\cite{he2020momentum} for the general tasks. He \etal leverage the memory bank to better train the model to exploit the similarity between a sample and the instances in the global memory bank~\cite{he2020momentum}. Wang et al. propose to use memory bank to facilitate hard negative instance mining across batches \cite{wang2020cross}. ECN~\cite{zhong2019invariance} leverages memory bank to enforce exemplar-invariance, camera-invariance and neighborhood-invariance over global training data (instead of local batch) for UDA person ReID. We introduce contrastive loss to the target instance memory bank to enable the joint optimization of positive pairs and negative pairs for a query/anchor sample over all the samples in the memory bank, which serves as our strong baseline.


	\begin{figure*}[t]
		\centering
		\includegraphics[width=0.90\linewidth]{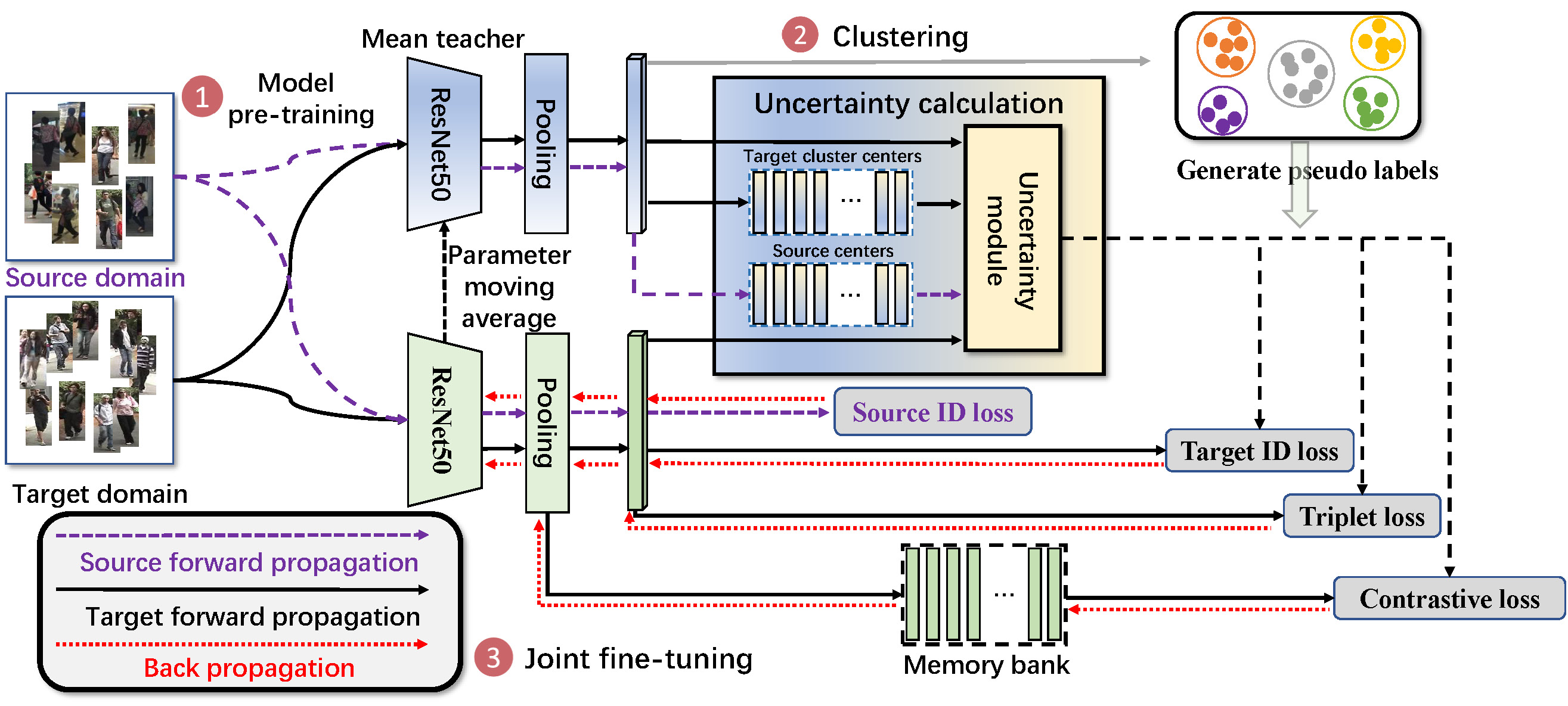}
		\caption{Overview of the proposed Uncertainty-guided Noise Resilient Network (UNRN) for UDA person ReID. We build our baseline framework with the mean teacher method (where the mean teacher model is a temporally moving average of weights of the student network) together with contrastive loss (supported by a memory bank). Our method belongs to clustering-based methods. In the model pre-training stage, we pre-train the network using source domain labeled data. In the clustering stage, we do clustering on the unlabeled target domain data using the more accurate features from the mean teacher model and assign pseudo labels based on the clustering results. Because of domain gap, some of the pseudo-labels are noisy/incorrect. In the joint fine-tuning stage, we propose to exploit the estimated uncertainty to evaluate the reliability of the pseudo-labels to alleviate the negative influence of the samples with error-prone pseudo-labels, by carefully incorporating the uncertainty to re-weight the contribution of samples in ID classification loss, triplet loss, and contrastive loss, respectively. Stage 2 and Stage 3 are performed alternatively.}
		\label{fig:fm}
	\end{figure*}

	\subsection{Uncertainty Estimation}

	Uncertainty modeling helps us to understand what a model does not know (or is not confident). In order to suppress the noise during the training, existing works~\cite{kendall2017uncertainties, chang2020data, zheng2020rectifying,zheng2019abstract} have explored uncertainty estimation from different aspects, such as the data-dependent uncertainty, model uncertainty, and the uncertainty on annotation. For fully supervised learning tasks with clean groudtruth labels, some works learn the data-dependent uncertainty in an end-to-end manner to alleviate the influence of observation noise of an input sample for better network optimization~\cite{kendall2017uncertainties, chang2020data}.
	Zheng \etal~\cite{zheng2020rectifying} use an extra auxiliary classifier to help estimate the uncertainty of the predicted pseudo labels for semantic segmentation. For clustering-based UDA person ReID, some of the pseudo labels generated by clustering are incorrect/noisy. However, the investigation on how to softly evaluate the reliability of clustering-generated pseudo labels is still underexplored. In this work, we explore the reliability estimation of the clustering-generated pseudo labels and alleviate the influence of noise-prone samples by re-weighting their contributions in ReID losses.

\section{\!\!\!\!Uncertainty-guided Noise Resilient Network}

Unsupervised domain adaptive person ReID aims at adapting the model trained on a labeled source domain dataset $\mathbb{D}_{s}\!\!=\!\!\left\{\!\!\left.\left(\boldsymbol{x}_{i}^{s}, {y}_{i}^{s}\right)\right|_{i=1} ^{N_{s}}\!\right\}$ \tcb{of $C_{s}$ identities} to an annotation-free target domain dataset $\mathbb{D}_{t}\!\!=\!\!\left\{\!\left.\boldsymbol{x}_{i}^{t}\right|_{i=1} ^{N_{t}}\right\}$. $N_{s}$ and $N_{t}$ denote the number of samples. $\boldsymbol{x}_{i}$ denotes a sample and ${y}_{i}$ denotes its groudtruth label.
Fig.~\ref{fig:fm} shows an overview of our proposed Uncertainty-guided Noise Resilient Network (UNRN) for UDA person ReID. We aim to address the negative influence of noise-prone pseudo labels during adaptation/fine-tuning under the clustering-based framework. We build a clustering-based strong baseline scheme \emph{SBase.} for UDA person ReID. On top of the strong baseline, we introduce an uncertainty calculation module to estimate the reliability of psedudo labels and incorporate the uncertainty into the ReID losses (ID classification loss (ID loss), triplet loss, and contrastive loss, respectively) to alleviate the influence of noise-prone pseudo labels. Our uncertainty-guided optimization design brings significant improvement on top of this strong baseline. We introduce the strong baseline in Section \ref{subsec:baseline} and elaborate on our proposed uncertainty-guided noise-resilient optimization designs in Section \ref{subsec:uncertainty}.






\subsection{Clustering-based Strong Baseline}
\label{subsec:baseline}

We build a clustering-based strong baseline scheme \emph{SBase.} for UDA person ReID. We follow the general pipeline of clustering-based UDA methods \cite{fan2018unsupervised, song2020unsupervised, jin2020global} which consists of three main stages, \ieno, model pre-training, clustering, and fine-tuning. As shown in Fig.~\ref{fig:fm}, we exploit the labeled source domain data for fine-tuning, incorporate contrastive loss, and leverage the simple yet effective mean teacher method to have a strong baseline. We will present the ablation study of each component in the experimental section.

In the model pre-training stage, we pre-train the network using source domain labeled data. In the clustering stage, we do clustering on the unlabeled target domain data using the more accurate features from the mean teacher model and generate pseudo labels based on the clustering results.

\noindent\textbf{Joint fine-tuning using source data.} In the fine-tuning stage~\cite{zhang2019self}, \tcb{most works} fine-tune the networks only using target domain pseudo labels. Here, we also re-use the valuable source domain data with reliable groudtruth labels. \tcb{For a source sample, we add ID classification loss, where maintained $C_{s}$ class centers are used as the classification weight vectors for classification.}


\noindent\textbf{Contrastive loss across memory bank.} For image retrieval task, to enable the pair similarity optimization over informative negative instances, Wang \etal propose a cross-batch memory mechanism that memorizes the feature embeddings of the past iterations to allow collecting sufficient hard negative pairs across the memory bank for network optimization~\cite{wang2020cross}. Motivated by this, for a \tcb{target domain} query sample $a$, we add contrastive loss to maximize the within-class similarity and minimize the between-class similarity across \tcb{the memory bank. Particularly, the memory bank consists of $N$ target domain instances (which is maintained similar to \cite{wang2020cross}) and $C_{s}$ source class center features (as the negative samples). Based on the pseudo labels of the target domain samples and the additional source class centers, we have $N_a^+$ positive samples and $N_a^- = N-N_a^+ + C_{s}$ negative samples. We optimize their similarities with respect to the query sample.} Following circle loss \cite{sun2020circle}, we use self-paced weighting to softly emphasize harder sample pairs by giving larger weights to them to get effective update.

\noindent\textbf{Mean teacher method.} Mean teacher is a method that temporally averages model weights over training steps, which tends to produce a more accurate model than using the final weights directly \cite{tarvainen2017mean}. As illustrated in Fig.~\ref{fig:fm}, there is no gradient back-propagation over the teacher model which just maintains a temporal moving average of the student model. This method is simple yet effective. We use the features from the teacher model to perform clustering and the final ReID inference.




\subsection{Uncertainty-guided Optimization}
\label{subsec:uncertainty}


\begin{figure}[!t]
	\centering
	\includegraphics[width=1.0\linewidth]{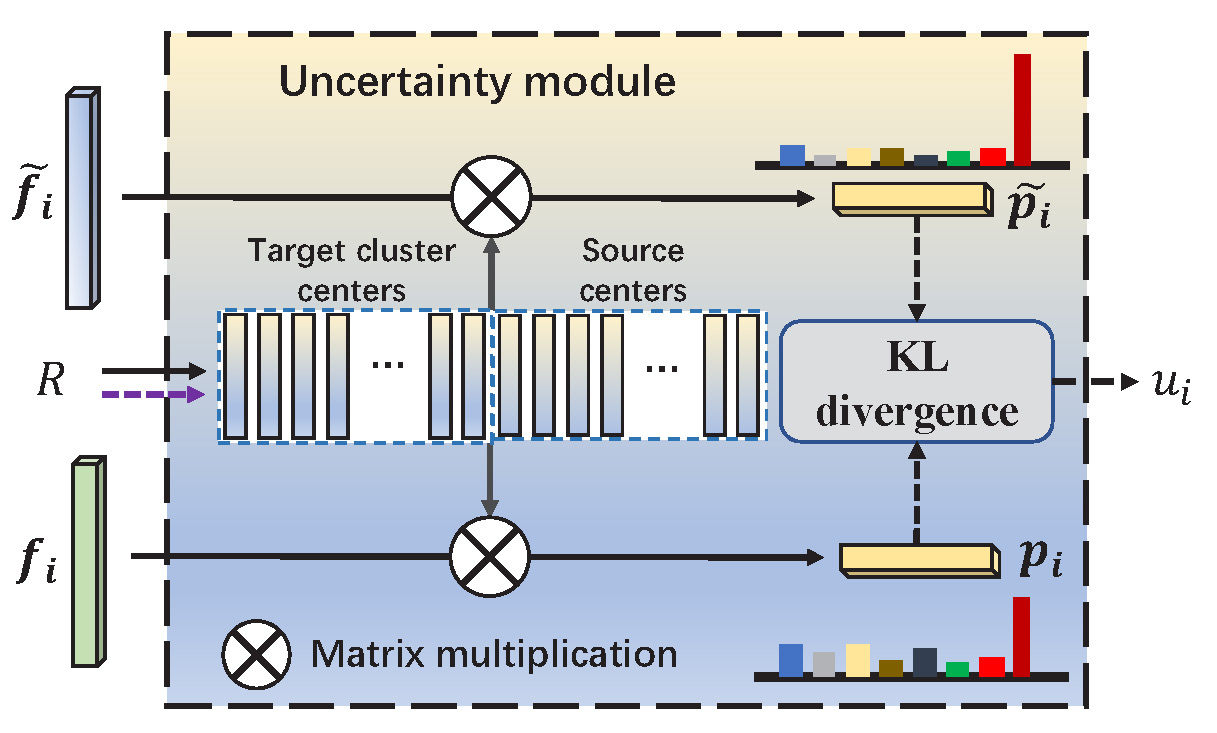}
	\caption{Uncertainty estimation module. For a sample $\boldsymbol{x}_i$ and its feature $\boldsymbol{\tilde{f}}_i$ from the mean teacher model and feature $\boldsymbol{f}_i$ from the student model, we estimate the uncertainty of its pseudo label by calculating the KL divergence of their soft multilabels.}
	\label{fig:fm2}
\end{figure}

The noisy pseudo labels would mislead the feature learning in the fine-tuning stage and hurt the adaptation performance. We aim to reduce the negative influence of noisy pseudo labels by evaluating the credibility of the pseudo label of each sample and suppress the contributions of samples with error-prone pseudo labels in the ReID losses.

\noindent{\textbf{Uncertainty estimation.}} Intuitively, the higher of the uncertainty the model has on its output of a sample, the lower of the reliability (larger of observation noise) of the output and it is more likely to have incorrect pseudo labels through clustering. We have observed in Fig.~\ref{fig:intro} that the samples with wrong pseudo labels in general have higher uncertainty values than those with correct pseudo labels. We leverage the uncertainty to softly assess the credibility of samples.

We estimate the uncertainty based on the output consistency between the mean teacher model and the student model. For a sample $\boldsymbol{x}_i$ of the target domain, we denote the extracted feature from the student model as $\boldsymbol{f}_i \in \mathbb{R}^{D}$ of $D$ dimensions, and the feature from the teacher model as $\boldsymbol{\tilde{f}}_i \in \mathbb{R}^{D}$. One straightforward solution to calculate the consistency between the mean teacher model and the student model is to calculate the distance (\egno, cosine distance) between the two features. However, this is clustering ignorance which does not capture/explore the global distribution of the target domain samples.

In MAR \cite{yu2019unsupervised}, they measure the class likelihood vector (\ieno, soft multilabel) of a person image with a set of reference persons (from an auxiliary domain). The inconsistency with the soft multilabel vector of an anchor sample is used to mine the hard negative sample that is visually similar to the anchor image but is of a different identity. Inspired by this, we propose to evaluate the uncertainty based on the inconsistency of the two features $\boldsymbol{f}_i$ and $\boldsymbol{\tilde{f}}_i$, by calculating the distance (\egno, KL distance, L1 distance) of their soft multilabels with respect to the same set of cluster centers. Particularly, as illustrated in Fig.~\ref{fig:fm2}, we use the class centers of source dataset and the cluster centers of the target domain data together to form the set of ``reference persons''. The soft multilabel agreement is analog to the voting by the ``reference persons'' to evaluate the relative consistency. Besides, through the comparison with the cluster centers of the target domain data, the soft multilabel captures some global information of the clustering centers which is related to pseudo label assignment. As an auxiliary domain, the source centers provide additional references.

Let ${R} =[R_t, R_s] \in \mathbb{R}^{K_r \times D}$ denote a matrix which stores the features of the $K_r$ ``reference persons'', with each column denoting the feature of a ``reference person''. $R_t \in \mathbb{R}^{K_t \times D}$ denotes the $K_t$ cluster centers of the target domain data, and $R_s \in \mathbb{R}^{K_s \times D}$ denotes the $K_s$ class centers of the source domain data (which are obtained from the weights of the fully-connected layer of the ID classifier). $K_r=K_t+K_s$.

For a feature $\boldsymbol f_i \in \mathbb{R}^{D}$ from the student model, we calculate the similarity to the $K_r$ ``reference persons'' and obtain the soft multilabel (likelihood vector) as:
\begin{equation}
\boldsymbol p_i = {\rm{Softmax}} \left( R \cdot \boldsymbol f_i \right),
\end{equation}
where Softmax($\cdot$) denotes softmax function which normalizes the similarity scores. Similarly, for the feature $\boldsymbol{\tilde{f}}_i$ from the mean teacher model, we obtain its soft multilabel as $\boldsymbol{\tilde{p}}_i$. We use KL divergence to measure the difference between the two probability distributions from the two models as the uncertainty $u_i$ of the sample $\boldsymbol x_i$:
\begin{equation}
      u_i = D_{KL}(\boldsymbol{\tilde{p}}_i || \boldsymbol p_i)= \sum_{k=1}^{K_r} \tilde{p}_{i,k} \log \frac{\tilde{p}_{i,k}}{p_{i,n}}.\\
\end{equation}

\noindent{\textbf{Optimization.}} We have observed that a sample with a wrong pseudo-label (through clustering), in general, has a higher uncertainty. Based on this observation, we propose to exploit the uncertainty to estimate the unreliability of the pseudo-label of a sample and use it to re-weight the contributions of samples in various ReID losses.
For a sample $\boldsymbol{x}_i$ with high uncertainty, we will reduce its contribution to the losses. Therefore, we could assign $\omega_i= 1/u_{i}$ as the credibility weight. To enable more stable training, we adopt the policy in~\cite{kendall2017uncertainties} and define $\omega_i = {\rm{exp}}({-u_i})$.
We incorporate the uncertainty-guided optimization in the classification loss, triplet loss, and contrastive loss, respectively.

For \emph{ID Classification loss}, we define the uncertainty-guided ID classification loss in a min-batch of $n_t$ target domain samples as
\begin{equation}
\mathcal{L}_{UID} =-\frac{1}{n_{t}} \sum_{i=1}^{n_{t}} \mathcal \omega_{i} \log p\left(\tilde{y}_{i} | \boldsymbol{x}_{i}\right),
\label{eq:9}
\end{equation}
where $p\left(\tilde{y}^t_{i} | \boldsymbol{x}^t_{i}\right)$ denotes the probability of being class $\tilde{y}^t_{i}$, where $\tilde{y}^t_{i}$ denotes the pseudo groudtruth class (based on the pseudo label assigned after clustering). For a sample with high uncertainty, a smaller weight is used to reduce its contribution to the overall loss to reduce its negative effect.

As a typical sample-pair similarity optimization, triplet loss is widely used in ReID to make the similarity between an anchor sample and a positive sample to be much larger than that between this anchor sample and negative sample. For the $j^{th}$ triplet of an anchor sample, a positive sample and a negative sample that correspond to three pseudo labels, we approximate the reliability of a sample pair, \egno, the positive sample pair, by a function of the two uncertainties as
\begin{equation}
\mathcal{\omega}_{ap}^j = \varphi(u_{a}^j, u_{p}^j),
\label{eq:omega}
\end{equation}
where $u_a^j$ and $u_p^j$ denote the estimated uncertainty for the anchor sample and positive sample in the $j^{th}$ triplet, respectively. For simplicity, we define the pair credibility as the average of two credibility as $\varphi(u_{a}^j, u_{p}^j) = \omega_a^j + \omega_p^j = {\rm{exp}}({-u_a^j}) + {\rm{exp}}({-u_p^j})$. Similarly, we get $\mathcal{\omega}_{an}^j$ for the negative sample pair.

For \emph{triplet loss}, we define the uncertainty-guided triplet loss in a min-batch of $n_{tr}$ triplets as
\small
\begin{equation}
    \mathcal{L}_{UTRI}\!=-\frac{1}{n_{tr}} \sum\limits_{j=1}^{n_{tr}} \log  \frac{\omega_{ap}^j \exp(s_{ap}^j)}{ \omega_{ap}^j \exp(s_{ap}^j)\!+\! \omega_{an}^j \exp(s_{an}^j)},
\end{equation}
\normalsize
where $s_{an}^j$ denotes the similarity for the $j^{th}$ positive sample pair. Mathematically, the lower credibility (higher uncertainty) of a sample pair, the smaller of a weight on the similarity and thus a smaller gradient in optimization, \ieno, contributing smaller to the optimization.


For \emph{contrastive loss}, given a query/anchor sample $\boldsymbol{x}_k$, we have $N_k^+$ positive samples and $N_k^-$ negative samples in the memory bank. For a batch of $n_t$ samples, by introducing the sample pair credibility weights, the uncertainty-guided contrastive loss is as


\small
\begin{equation}
\mathcal{L}_{\text {UCT}}\!=\!\frac{1}{n_t} {\sum\limits_{k=1}^{n_t}}\log\!\!\left[\!\! 1+{\sum\limits_{j=1}^{N_k^-} \omega_{kj}^- {\rm{exp}}(s_{kj}^-) \sum\limits_{i=1}^{N_k^+} \omega_{ki}^+ {\rm{exp}}(-s_{ki}^+) }\right], 
\label{eq:UCT}
\end{equation}
\normalsize
where $s_{kj}^-$ denotes the similarity between the query sample $\boldsymbol{x}_k$ and the $j^{th}$ negative sample, and $\omega_{kj}^-$ denotes the approximated reliability of the sample pair (see (\ref{eq:omega})). The lower credibility of a sample pair, the smaller the gradient and the contribution of this pair to the optimization.  \tcb{Note that similar to our strong baseline, we also use self-paced weighting ~\cite{sun2020circle} to softly emphasize harder sample pairs (whose similarity score deviates far from the optimum) by giving larger weights to them to get effective update. For simplicity, we do not present it in (\ref{eq:UCT}), where $s_{kj}^-$, $s_{ki}^+$ denote the similarities that already re-weighted. }

The total loss for the target domain data in the fine-tuning stage could be formulated as:
\begin{equation}
\mathcal L_{target} = \mathcal L_{UID} + \lambda_{tri} \mathcal L_{UTRI} + \lambda_{ct} \mathcal L_{UCT} + \lambda_{reg} \mathcal L_{reg},
\end{equation}
where $L_{reg} = \frac{1}{n_t} {\sum_{i=1}^{n_t}} u_i$ is a regularization loss which prevents large uncertainty (\ieno, small credibility which could reduce the first three losses) all the time. $\lambda_{tri}$, $\lambda_{ct}$, and $\lambda_{reg}$ are weighting factors.

	\section{Experiments}

	\begin{table*}[t!]
		\footnotesize
		\centering
		\caption{Performance (\%) comparison with the state-of-the-art methods for UDA person ReID on the datasets of DukeMTMC-reID, Market-1501 and MSMT17. We mark the results of the second best by \underline{underline} and the best results by \textbf{bold} text.
		}
		\vspace{-3mm}
		\begin{center}
			\begin{tabular}{|P{6.0cm}|C{1.0cm}C{1.0cm}C{1.0cm}C{1.0cm}|C{1.0cm}C{1.0cm}C{1.0cm}C{1.0cm}|}
				\hline
				\multicolumn{1}{|c|}{\multirow{2}{*}{Methods}} & \multicolumn{4}{c|}{DukeMTMC$\to$Market1501} & \multicolumn{4}{c|}{Market1501$\to$DukeMTMC} \\
				\cline{2-9}
				\multicolumn{1}{|c|}{} & mAP & R1 & R5 & R10 & mAP & R1 & R5 & R10 \\
				\hline
				ATNet~\cite{Liu2019cvpr}(CVPR'19) & 25.6& 55.7& 73.2& 79.4& 24.9 &45.1 &59.5& 64.2\\
				SPGAN+LMP~\cite{DengWeijian2018cvpr}(CVPR'18) &26.7 &57.7 &75.8 &82.4& 26.2&46.4 &62.3 &68.0 \\
				CFSM~\cite{chang2018disjoint} (AAAI'19) & 28.3 & 61.2 & - & - & 27.3 & 49.8 &- & -  \\
				BUC~\cite{lin2019aBottom} (AAAI'19) & 38.3 & 66.2 & 79.6 & 84.5 & 27.5 & 47.4 & 62.6 & 68.4 \\
				ECN~\cite{zhong2019invariance} (CVPR'19) & 43.0 & 75.1 & 87.6 & 91.6 & 40.4 & 63.3 & 75.8 & 80.4 \\
				UCDA~\cite{qi2019novel} (ICCV'19) & 30.9 & 60.4 & - & - & 31.0 & 47.7 & - & - \\
				PDA-Net~\cite{li2019cross} (ICCV'19) & 47.6 & 75.2 & 86.3 & 90.2 & 45.1 & 63.2 & 77.0 & 82.5 \\
				PCB-PAST~\cite{zhang2019self} (ICCV'19) & 54.6 & 78.4 & - & - & 54.3 & 72.4 & - & - \\
				SSG~\cite{yang2019selfsimilarity} (ICCV'19) & 58.3 & 80.0 & 90.0 & 92.4 & 53.4 & 73.0 & 80.6 & 83.2 \\
				ACT~\cite{Yang2019Asymmetric} (AAAI'20) & 60.6 & 80.5 & - & - & 54.5 & 72.4 & - & - \\
				MPLP~\cite{WANG2020cvpr1} (CVPR'20) & 60.4 &84.4 &92.8& 95.0 & 51.4&72.4 &82.9& 85.0  \\
				DAAM~\cite{Huang2020aaai} (AAAI'20)& {67.8} & {86.4} & {-} & {-} & {63.9} & {77.6} & {-} & {-} \\ 
				AD-Cluster~\cite{zhai2020adcluster} (CVPR'20)& {68.3} & {86.7} & {94.4} & {96.5} & {54.1} & {72.6} & {82.5} & {85.5} \\
				MMT~\cite{ge2020mutual} (ICLR'20) & {71.2} & {87.7} & {94.9} & {96.9} & {65.1} & {78.0} & \underline{88.8} & \underline{92.5} \\
				NRMT~\cite{zhao2020unsupervised}(ECCV'20) & 71.7& 87.8& 94.6& 96.5& 62.2& 77.8& 86.9& 89.5 \\
				B-SNR+GDS-H~\cite{jin2020global}(ECCV'20) & 72.5 & 89.3 & - & - & 59.7 & 76.7 &- &-\\
				MEB-Net~\cite{zhai2020multiple}(ECCV'20) & \underline{76.0} &\underline{89.9}& \underline{96.0}& \underline{97.5}& \underline{66.1}& \underline{79.6} & 88.3& 92.2 \\
				\hline

				UNRN (Ours) & \textbf{78.1} & \textbf{91.9} & \textbf{96.1} & \textbf{97.8} & \textbf{69.1} & \textbf{82.0} & \textbf{90.7} & \textbf{93.5} \\

				\hline

			\end{tabular}\\

			\begin{tabular}{|P{6.0cm}|C{1.0cm}C{1.0cm}C{1.0cm}C{1.0cm}|C{1.0cm}C{1.0cm}C{1.0cm}C{1.0cm}|}
				\hline
				\multicolumn{1}{|c|}{\multirow{2}{*}{Methods}} & \multicolumn{4}{c|}{Marke1501$\to$MSMT17} & \multicolumn{4}{c|}{DukeMTMC$\to$MSMT17} \\
				\cline{2-9}
				\multicolumn{1}{|c|}{} & mAP & R1 & R5 & R10 & mAP & R1 & R5 & R10 \\
				\hline
				ECN~\cite{zhong2019invariance} (CVPR'19) & 8.5 & 25.3 & 36.3 & 42.1 & 10.2 & 30.2 & 41.5 & 46.8 \\
				SSG~\cite{yang2019selfsimilarity} (ICCV'19) & 13.2 & 31.6 &- & 49.6 & 13.3 & 32.2 & - & 51.2 \\
				DAAM~\cite{Huang2020aaai} (AAAI'20)& {20.8} & { 44.5} & {-} & {-} & { 21.6} & { 46.7} & {-} & {-} \\
				NRMT~\cite{zhao2020unsupervised}(ECCV'20) & 19.8& 43.7& 56.5 &62.2& 20.6 &45.2& 57.8& 63.3 \\
				MMT~\cite{ge2020mutual} (ICLR'20) & \underline{22.9} & \underline{49.2} & \underline{63.1} & \underline{68.8} & \underline{23.3} & \underline{50.1} & \underline{63.9} & \underline{69.8} \\
				\hline
				 UNRN (Ours) & \textbf{25.3} & \textbf{52.4} & \textbf{64.7} & \textbf{69.7} & \textbf{26.2} & \textbf{54.9} & \textbf{67.3} & \textbf{70.6} \\

				\hline

				\hline
			\end{tabular}
		\end{center}
		\label{tab:sota}
	\end{table*}

	\subsection{Datasets and Evaluation Metrics}

	We evaluate our methods using three person ReID datasets, including DukeMTMC-reID (Duke) ~\cite{dukemtmc}, Market-1501 (Market) ~\cite{market} and MSMT17~\cite{wei2018person}.
	DukeMTMC-reID~\cite{dukemtmc} has 36,411 images, where 702 identities are used for training and 702 identities for testing.
	Market-1501~\cite{market} contains 12,936 images of 751 identities for training and 19,281 images of 750 identities for testing.
	MSMT17~\cite{wei2018person} contains 126,441 images of 4,101 identities, where 1,041 identities and 3060 identities are used for training and testing respectively.

	We adopt mean average precision (mAP) and CMC Rank-1/5/10 (R1/R5/R10) accuracy for evaluation.


	\subsection{Implementation Details}
	\label{sec:imp}

	We use ResNet50 pretrained on ImageNet as our backbone networks. As ~\cite{luo2019bag}, we perform data agumentation of randomly erasing, cropping, and flipping. For source pre-training, each mini-batch contains 64 images of 4 identities. For our fine-tuning stage, when the source data is also used, each mini-batch contains 64 source-domain images of 4 identities and 64 target-domain images of 4 pseudo identities, where there are 16 images for each identity. All images are resized to 256$\times$128. Similar to \cite{ge2020mutual,yang2019selfsimilarity}, we use the clustering algorithm of DBSCAN. For DBSCAN, the maximum distance between neighbors is set to $eps=0.6$ and the minimal number of neighbors for a dense point is set to 4. ADAM optimizer is adopted. The initial learning rate is set to 0.00035. We set the weighting factors $\lambda_{tri}=1$,  $\lambda_{ct}=0.05$, and $\lambda_{reg}=1$, where we determine them simply by making the several losses on the same order of magnitude.

	\subsection{Comparison with the State-of-the-arts}

	We compare our proposed UNRN with the state-of-the-art methods on four domain adaptation settings in Tab.~\ref{tab:sota}. Our UNRN significantly outperforms the second best UDA methods by $2.1\%$, $3.0\%$, $2.4\%$ and $2.9\%$, in mAP accuracy, for Duke$\to$Market, Market$\to$Duke, Market$\to$MSMT, and Duke$\to$MSMT, respectively.
	SSG~\cite{yang2019selfsimilarity} performs multiple clustering on both global body and local body parts. DAAM~\cite{Huang2020aaai} introduces an attention module and incorporates domain alignment constraints. MMT~\cite{ge2020mutual} use two networks (four models) and MEB-Net~\cite{Zhai2020} use three networks (six models) to perform mutual mean teacher training, which have high computation complexity in training. Our UNRN uses only one network (two models) in training but still significantly outperforms the best-performing MEB-Net ~\cite{Zhai2020}.

	\subsection{Ablation Studies}


	\begin{table}[h!]
		\caption{Ablation studies on the effectiveness of components in our proposed UNRN on Market and Duke. \textbf{Source}: source data is also used in fine-tuning stage with ID classification loss. \textbf{Contrastive loss (CT)}: pair similarity optimization across the memory bank. \textbf{Mean teacher}: use mean teacher method where a temporally moving averaged model is taken as the mean teacher. \textbf{ID}: target domain identity classification loss. \textbf{UID}: ID loss with \emph{uncertainty}.  \textbf{TRI}: target domain triplet loss. \textbf{UTRI}: target domain triplet loss with \emph{uncertainty}. \textbf{UCT}: contrastive loss with \emph{uncertainty}.}
		\centering
		\footnotesize
		\vspace{-5pt}
		\label{tab:ablation1}
		\begin{center}

			\begin{tabular}{P{3.4cm}|C{0.7cm}C{0.7cm}|C{0.7cm}C{0.85cm}}

				\hline
				\multicolumn{1}{c|}{\multirow{2}{*}{Methods}} & \multicolumn{2}{c|}{Duke$\to$Market}&\multicolumn{2}{c}{Market$\to$Duke}   \\
				\cline{2-5}
				\multicolumn{1}{c|}{}& mAP & R1 & mAP & R1 \\
				\hline \hline
				Supervised learning    &85.7&94.1&75.8&86.2\\
				\hline
				\hline
				Model pretraining      &32.9&62.6&35.2&53.3\\
				\hline
				Baseline               &68.2&87.9&60.4&75.9\\
				+Source                &70.4&88.3&61.3&76.4\\
				+Contrastive loss      &72.1&88.7&62.1&77.6\\
				+Mean teacher (SBase.) &75.4&89.8&64.8&79.7\\
				\hline
				SBase.~(ID+TRI+CT)     &75.4&89.8&64.8&79.7\\
				SBase.~w/~UID          &77.3&91.2&68.2&81.3\\
				SBase.~w/~UTRI         &76.1&90.9&66.3&81.0\\
				SBase.~w/~UID+UTRI     &77.8&91.5&68.9&81.7\\
				SBase.~w/~UID+UTRI+UCT &78.1&91.9&69.1&82.0\\
				\hline
			\end{tabular}


		\end{center}
		\vspace{-15pt}
	\end{table}

	\noindent\textbf{Effectiveness of components in our strong baseline.} We build our basic baseline \emph{Baseline} following the commonly used baselines in UDA methods~\cite{ge2020mutual,song2020unsupervised,jin2020global}, where in the fine-tuning stage, the identity classification loss and triplet loss are used to fine-tune the network based on the pseudo labels for the target domain data. On top of \emph{Baseline}, we add three components (as described in Section \ref{subsec:baseline}. Tab.~\ref{tab:ablation1} shows that each component brings additional significant gain and finally we have a strong baseline \emph{SBase.}.

	\noindent\textbf{Effectiveness of our uncertainty-guided optimization.} We validate the effectiveness of our proposed design on top of a strong baseline \emph{SBase.}. In general, the stronger of a baseline, the harder one can achieve gains since the cases easy to address are mostly handled by the strong baseline. Once the new design is complementary to the strong baseline, it is valuable to advance the development of techniques.

	In the strong baseline \emph{SBase.}, for the target domain samples, identity classification loss (ID), triplet loss (TRI), and contrastive loss (CT) are used for supervision based on pseudo labels. To alleviate the negative influence of noisy/wrong pseudo labels, we exploit uncertainty to re-weight the contributions of samples.
	Tab.~\ref{tab:ablation1} shows the comparisons. When we replace ID loss by UID loss, which is the ID loss with uncertainty, the mAP accuracy is significantly improved by 1.9\% and 3.4\% for Duke$\to$Market and Market$\to$Duke, respectively. When we replace triplet (TRI) loss by UTRI loss, similar improvements are observed. We have our final scheme \emph{UNRN} when the uncertainty-guided optimization is applied to all the three losses and it achieves \textbf{2.7\%} and \textbf{4.3\%} improvement in mAP accuracy over \emph{SBase.} for Duke$\to$Market and Market$\to$Duke, respectively.


	\subsection{Design Choices}

	\noindent\textbf{Influence of different designs in uncertainty estimation.} We have discussed the estimation of uncertainty in Section \ref{subsec:uncertainty}. There are some design choices and Tab.~\ref{tab:ablation2} shows the comparisons. \textbf{1)} \emph{UNRN-feat. consist.} denotes that we estimate the uncertainty based on the distance of the features $\boldsymbol{\tilde{f}_i}$ and $\boldsymbol{{f}_i}$ instead of the distance of derived soft multilabels. We can see that the gain over \emph{SBase.} is negligible due to the poor estimation of uncertainty. In contrast, using the consistency between soft multilabels (\ieno \emph{UNRN-$R$ (Ours)}) captures global information of the target cluster centers and brings significant improvement. \textbf{2)} When we estimate the uncertainty, we leverage a set of ``reference persons'' to get the soft multilabels. \emph{UNRN-$R_t$} denotes the scheme when only the target cluster centers (see Fig.~\ref{fig:fm2}) are used as ``reference persons''.  \emph{UNRN-$R_s$} denotes the scheme when only the source centers are used as ``reference persons''. \emph{UNRN-$R$} denotes both are used as ``reference persons''. We can see that the performance of \emph{UNRN-$R_s$} is very similar to that of \emph{SBase.}, where the source centers only cannot provide the clustering information of target domain data and is helpless to estimate the reliability of pseudo labels. \emph{UNRN-$R_t$} outperforms \emph{SBase.} significantly, which captures the target domain global clustering information that is helpful to estimate the reliability of pseudo labels. Interestingly, \emph{UNRN-$R$} which jointly considers the target cluster centers and source centers provides the best performance. That may
    be because the source centers provide more references which enables the soft multilabels more informative.      
	
	\noindent\textbf{Influence of regularization loss $\mathcal{L}_{reg}$.} The regularization loss $\mathcal{L}_{reg}$ prevents larger uncertainty all the time. As shown in Tab.~\ref{tab:ablation2}, our final scheme \emph{UNRN-$R$} outperforms \emph{UNRN w/o $\mathcal{L}_{reg}$} by 0.6\% and 1.1\% in mAP accuracy for Duke$\to$Market and Market$\to$Duke, respectively.


	\begin{table}[t]
		\caption{Influence of different designs in uncertainties estimation, and the influence of regularization loss $\mathcal{L}_{reg}$ on performance under our framework.}
		\centering
		\small
		\vspace{-2mm}
		
		\label{tab:ablation2}
		\begin{center}
			
			\begin{tabular}{P{3.6cm}|C{0.7cm}C{0.7cm}|C{0.7cm}C{0.85cm}}
				\hline
				\multicolumn{1}{c|}{\multirow{2}{*}{Methods}} & \multicolumn{2}{c}{Duke$\to$Market}&\multicolumn{2}{c}{Market$\to$Duke}   \\
				\cline{2-5}
				\multicolumn{1}{c|}{}& mAP & R1 & mAP& R1 \\ 
				\hline \hline
				SBase.                       &75.4&89.8&64.8&79.7\\
				\hline
				UNRN w/o $\mathcal{L}_{reg}$ &77.5&91.4&68.0&81.4\\
				\hline
				UNRN-feat. consist.          &76.5&91.0&66.7&80.9 \\
				UNRN-$R_s$                   &76.0&91.3&66.6&81.0 \\ 
				UNRN-$R_t$                   &77.8&91.7&68.3&81.7 \\ 
				UNRN-$R$ (Ours)              &78.1&91.9&69.1&82.0 \\ 
				\hline
			\end{tabular}
		\end{center}
		\vspace{-10pt}
	\end{table}
	
	\noindent\textbf{Influence of size of memory bank.} We use a queue to maintain \tcb{$N$ target domain instances in the} memory bank. Tab.~\ref{tab:memory-size} shows that as the queue length $N$ increases, the performance increases but saturates when the size is around 8192. 	
	
	\begin{table}[t]
	\caption{Influence of the number ($N$) of target instances in the memory bank. We study this on top of baseline scheme \emph{Baseline+Source+Contrastive loss} (see Tab.~\ref{tab:ablation1}). ``All'' denotes the size is equal to the size of target training dataset.}
	\label{tab:memory-size}
	\vspace{-8pt}
	\begin{center}
	\small
	\begin{tabular}{c|c|c|c|c|c}
    \hline
    \multicolumn{1}{c|}{\multirow{2}{*}{Size of memory bank}} & \multicolumn{5}{c}{Market$\to$Duke}\\	\cline{2-6}
    \multicolumn{1}{c|}{}& 0    & 1024 & 4096 & 8192 & All  \\ 
    \hline
    \hline
    mAP                 & 62.8 & 63.4 & 63.9 & 64.8 & 64.5 \\ \hline
    R1              & 77.9 & 78.9 & 79.3 & 79.7 & 79.3 \\ \hline
    \end{tabular}
    \end{center}
    \end{table}
    \vspace{-8pt}

	\section{Conclusion}
	
	In this paper, for clustering-based UDA person ReID, we aim to alleviate the negative influence of wrong/noisy labels during the adaptation. We have observed that a sample with a wrong pseudo-label through clustering in general has a weaker consistency between the output of the mean teacher model and the student model. Based on this finding, we propose to exploit the uncertainty (measured by consistency levels) to evaluate the reliability of the pseudo-label of a sample and incorporate the uncertainty to re-weight its contribution within various ReID losses, including the ID classification loss per sample, the triplet loss, and the contrastive loss. Our uncertainty-guided optimization brings significant improvement over our strong baseline and our scheme achieves the state-of-the-art performance on benchmark datasets.
	
	\section{Acknowledgments}
	
	This work was in part supported by the National Key R$\&$D Program of China under Grand 2020AAA0105702, National Natural Science Foundation of China (NSFC) under Grants U19B2038 and 61620106009.
	
	\section{Ethical Impact}
	Our method is proposed to help match/identify different persons across images, which can facilitate the development of smart retail systems in the future. When the person re-ID system is used to identify the pedestrian, it may cause a violation of human privacy. Therefore, governments and officials need to carefully formulate strict regulations and laws to ensure the legal use of ReID technology and strictly protect the data.
	
	
	\bibliographystyle{aaai}
	\bibliography{references}
	
\end{document}